\documentclass{article}

\usepackage[dblblindworkshop,final]{neurips_2026}
\workshoptitle{AISciK}
\usepackage[utf8]{inputenc}
\usepackage[T1]{fontenc}
\usepackage{amsmath}
\usepackage{graphicx}
\usepackage{hyperref}
\usepackage{url}
\usepackage{booktabs}
\usepackage{amsfonts}
\usepackage{nicefrac}
\usepackage{microtype}
\usepackage{xcolor}

\title{Measurable Does Not Imply Transportable: A Prospective Audit of Model Explanations in Multi-Year Field Trials}

\author{%
  {\normalfont\bfseries\normalsize
  Rayhaneh Shabani Nia \hspace{0.8em}
  Ali Karkehabadi}\\[5pt]
  {\normalfont\small
  University of California, Davis}\\[4pt]
  {\normalfont\small
  \texttt{\{rshabaninia,akarkehabadi\}@ucdavis.edu}}%
}

\begin{document}

\maketitle

\begin{abstract}
Model explanations are increasingly used as scientific evidence, yet it remains unclear what they license scientists to conclude beyond the data on which they were constructed. When a fine-grained explanation fails to replicate, the failure may be attributed to noisy measurement of its target, raising a fundamental question: does reliable measurement imply a reproducible explanation? We test this implication directly. We formalize \emph{prospective explanation transportability}: the extent to which the structure of an explanation learned from past data is recovered in a genuinely future context. This property is distinct from both feature importance and measurement reliability. Using multi-year Genomes-to-Fields maize trials, in which each new year is a real future context rather than a random resample, we evaluate 16 agronomically interpretable environmental concepts across five strictly forward splits and one prospectively locked test year. Coarse environment-level explanations transport reliably (median cosine $0.83$); their locked 2024 value, $0.811$, fell within a prespecified $80\%$ bootstrap percentile interval. For grain yield, genotype-conditioned explanations have a historical median cosine of $0.30$ and range from $0.006$ to $0.822$ across years. Reliability does not order these outcomes: the year with the highest split-half reliability had the lowest explanation transport. Within-context resampling further showed that G$\times$E explanations were reproducible within a fixed context ($0.77$--$0.95$) while exhibiting substantially lower prospective alignment across the 2020--2023 transitions ($0.01$--$0.36$). A second phenotype transports strongly at both resolutions, and non-temporal validation inflates apparent interaction-level transport by $+0.22$. These results show that target reliability and within-context explanation reproducibility do not by themselves establish when a model explanation can support a reproducible scientific claim.

\end{abstract}

\section{Introduction}

Interpretability is increasingly used as a bridge between predictive modeling and
scientific discovery. Feature-attribution methods can identify variables that
influence a model's predictions \citep{lundberg2017unified}, and a growing body
of work has emphasized that such explanations should be evaluated rather than
accepted solely on visual or intuitive grounds
\citep{doshivelez2017rigorous, adebayo2018sanity, hooker2019benchmark,
krishna2024disagreement}. Yet most validation asks whether an explanation is
faithful to a fitted model, stable under perturbation, or predictive under
held-out data. Scientific use asks a different question. If an attribution is
interpreted as evidence for a reproducible relationship, the relevant question
is whether that explanation survives a genuinely new scientific context. This
need is especially acute when model outputs are used to support scientific
claims: subtle evaluation failures, including leakage across temporal or grouped
structure, can produce overly optimistic conclusions in ML-based science
\citep{kapoor2023leakage}. This distinction becomes especially important when the scientific claim is
conditioned on additional context. A population-level association and a
subgroup-specific interaction may be derived from the same measurements, but
they need not have the same evidential support. Moreover, poor replication of a
fine-grained explanation can have at least two very different causes: the
underlying scientific target may be measured unreliably, or the target may be
measurable while its explanation changes across contexts. Predictive accuracy,
feature importance, and measurement reliability therefore do not by themselves
establish what we call \emph{prospective explanation transportability}: the
extent to which the structure of an explanation constructed from past data is
recovered in a future context. We evaluate these quantities separately; the
audit does not identify which mechanism operates in a given failure of
replication. Prior work uses changes in explanation distributions to detect
distribution shift \citep{mougan2022explanation}; our question is different:
whether an explanation itself survives prospective evaluation as a scientific
claim. Genotype $\times$ environment (G$\times$E) modeling provides a particularly
useful setting for studying this problem. Modern genomic prediction models
combine genetic and environmental information and increasingly expose
interpretable environmental drivers through attribution mechanisms
\citep{yu2025exgep}. However, predictive generalization and
explanation generalization are different objects. An environmental feature can
remain useful for predicting yield while the scientific interpretation of its
effect changes when the claim is conditioned on genotype. Multi-year field
trials provide a natural way to test this distinction because each new year
constitutes a future environmental context rather than another random sample
from the same temporal mixture, paralleling the motivation for temporal-shift
evaluation in Wild-Time \citep{yao2022wildtime}.

We introduce a prospective explanation audit for this setting. Using multi-year
Genomes-to-Fields maize trials, we organize environmental measurements into 16
agronomically interpretable concepts, in the broader spirit of concept-based
interpretability \citep{kim2018tcav}, while using prespecified observed
covariates rather than learned concept activation vectors. We study explanations
at two primary scientific resolutions. The coarse target is mean yield within an
environment; the finer target is an environment-specific genomic-response slope
defined along a reference-fitted genomic PC1 axis. For each of five historical
evaluations, explanations are constructed from past years and compared only with
the subsequent year. We quantify their alignment using cosine similarity and
separately estimate split-half reliability of the genotype-dependent target,
allowing measurement precision to be distinguished from future explanation
transport. The resulting historical analysis is then followed by a locked 2024
evaluation with a prespecified main-effect forecast and bootstrap percentile interval. Our primary result is a pronounced difference in prospective behavior for grain
yield. Across the five historical future-year evaluations, environment-level
explanations achieve a median cosine similarity of $0.831$, whereas
genotype-conditioned G$\times$E explanations achieve $0.295$ and vary from
near-zero to strong transport across years. Reliability does not order these
outcomes: in 2019, modest split-half reliability ($0.352$) coexisted with
strong transport ($0.822$), whereas in 2020 reliability rose to $0.798$ while
transport fell to $0.006$. Across the 2020--2023 transitions, within-context
G$\times$E explanation reproducibility was $0.77$--$0.95$, while prospective
cross-year alignment was only $0.01$--$0.36$. In the locked 2024 evaluation,
the main-effect cosine was $0.811$, inside the prespecified $80\%$ bootstrap percentile
interval $[0.755, 0.942]$ around the frozen forecast of $0.831$; the
corresponding G$\times$E cosine was $0.271$. We next test several alternative explanations for this gap. Conventional
non-temporal validation also changes the apparent conclusion: random environment
splits yield a G$\times$E cosine of $0.512$, compared with $0.295$ under strict
forward evaluation, while the coarse main effect does not show the same
optimism. The main-to-interaction ordering additionally persists under an
independent Ridge coefficient-based explanation model, arguing against an
explanation specific to XGBoost/SHAP. Importantly, however, fine-grained
conditioning does not preclude transport: for Grain Moisture, environment-level
and genotype-conditioned explanations both transport strongly, with median
cosines of $0.907$ and $0.859$, respectively.

Our contributions are threefold:

\begin{itemize}
    \item We formulate prospective explanation transportability as a validation
    target for interpretability-driven scientific discovery and distinguish it
    from model-derived importance and measurement reliability.

    \item We show that reliable measurement and high within-context
    reproducibility do not imply that a fine-grained explanation will reproduce
    in a future context.

    \item We show that the apparent validity of fine-grained explanations
    depends on the phenotype and on the validation protocol: non-temporal
    splitting selectively inflates their apparent transport, whereas a second
    phenotype transports strongly at both resolutions.
\end{itemize}

\section{Related Work}

\paragraph{Interpretable genotype--environment modeling.}
Modern genomic prediction increasingly combines genomic and environmental
information to model genotype $\times$ environment (G$\times$E) variation. The
Genomes-to-Fields initiative provides a large multi-year resource for this
setting and supports analyses of environmental covariates and G$\times$E
structure in North American maize \citep{lopezcruz2023g2f}. Recent explainable
and attention-based G$\times$E models also aim to expose environmental drivers
of maize performance \citep{yu2025exgep}. We ask a narrower
question: whether an explanation recovered from historical environments
reproduces in a future year.

\paragraph{Evaluating explanations beyond plausibility.}
SHAP provides an additive framework for attributing a model's prediction to its
input features \citep{lundberg2017unified}. A broad interpretability literature
argues that explanations should be evaluated rather than accepted for their
visual or domain plausibility alone. Sanity checks test whether attributions
respond to changes in model parameters or labels \citep{adebayo2018sanity}, and
perturbation or retraining evaluations assess whether selected features carry
predictive information \citep{hooker2019benchmark}. Explanation methods can
also disagree for the same prediction \citep{krishna2024disagreement},
while feature decorrelation has been explored for saliency-based
interpretability \citep{karkehabadi2026saliencydecor}.
Saliency-guided learning has also been studied in image models
\citep{karkehabadi2024smoot,karkehabadi2024hlgm} and genomic sequence
classification \citep{shabaninia2026attngen}. These approaches address
faithfulness, sensitivity, or robustness relative to a model
or observed distribution. We instead ask whether the structure of an
explanation is recovered in a naturally occurring future context.

\paragraph{Concept-based explanations.}
Concept-based methods seek explanations in terms that domain experts can
interpret and evaluate. TCAV, for example, represents a user-defined concept by
a direction learned from examples in a model's latent space and measures the
sensitivity of predictions to that direction \citep{kim2018tcav}. Our analysis
shares this aim at an agronomically meaningful level, but does not learn concept
activation vectors: the 16 concepts are prespecified, observed environmental
covariates. We test whether these domain-defined concept-level explanations
recur across years.

\paragraph{Distribution shift and prospective evaluation.}
Explanation Shift models changes in explanation distributions as an indicator
of predictive-performance change \citep{mougan2022explanation}. Our question is
narrower: we do not use explanations to detect or diagnose a shift, but ask
whether an explanation itself remains supported in a later context. Temporal
benchmarks such as Wild-Time motivate strictly chronological evaluation rather
than random resampling \citep{yao2022wildtime}, while work on ML-based science
has documented how leakage and inappropriate evaluation can produce
overoptimistic scientific conclusions \citep{kapoor2023leakage}. We use this
prospective framing while separately assessing measurement reliability,
phenotype, and validation protocol.

\section{Methodology}
\label{sec:method}

Our analysis separates three quantities: \emph{model-derived explanation},
\emph{measurement reliability} of the underlying scientific target, and
\emph{prospective transportability} of the resulting explanation. The primary
analysis was frozen before the final 2024 scientific evaluation. Analyses added
afterward are reported as secondary/post-hoc tests of the primary result.

\subsection{Data and prospective design}
\label{sec:method_data}

We use the Genomes to Fields (G2F) maize multi-environment resource
\citep{lopezcruz2023g2f}, which contains grain yield, environmental
measurements, and genomic marker information collected across multiple years
and locations. The processed historical analysis contains 107,399
hybrid--environment observations, approximately 270 year--location
environments, and 2,425 SNP markers. The final 2024 evaluation contains 9,486
observations from 1,063 hybrids across 22 eligible environments. The scientific unit for prospective comparison is an \emph{environment},
defined by year and location, rather than an individual plot. Plot-level
replication improves estimation within an environment but does not create an
independent future environmental context.

Historical evaluation is strictly forward in time:
\[
\begin{aligned}
2014\text{--}2018 &\rightarrow 2019, &
2015\text{--}2019 &\rightarrow 2020,\\
2016\text{--}2020 &\rightarrow 2021, &
2017\text{--}2021 &\rightarrow 2022,\\
2018\text{--}2022 &\rightarrow 2023.
\end{aligned}
\]
For every split, all representation and explanation quantities are estimated
from the historical reference period and evaluated on the subsequent target
year. The final prospective evaluation uses
\(2019\text{--}2023 \rightarrow 2024\). This design prevents a future target
environment from contributing to the reference explanation against which it is
evaluated.

\subsection{Scientific explanation targets}
\label{sec:method_targets}

Environmental measurements are represented by a frozen set of 16 agronomically
interpretable concepts summarizing heat, water availability, atmospheric
dryness, thermal accumulation, soil-water context, and canopy conditions. The
same concept space is used for both explanation resolutions.

\paragraph{Environment-level main effect.}
The coarse target is average yield in environment \(e\):
\[
\bar{Y}_{e}
=
\frac{1}{n_e}
\sum_{i\in e} Y_{ie},
\]
where \(Y_{ie}\) is the observed yield of hybrid \(i\) in environment \(e\).

\paragraph{Genotype-dependent response.}
The finer target measures how yield changes along a continuous genomic direction
within each environment. Genomic PCA is fit using the historical reference
data, and hybrids in the target data are projected using the frozen reference
transformation. Let \(z_i\) denote the resulting PC1 score. Within each
environment we estimate
\[
Y_{ie}
=
\alpha_e
+
\beta_{G,e} z_i
+
\epsilon_{ie},
\]
where \(\beta_{G,e}\) is the environment-specific genomic-response slope. The
vector of \(\beta_{G,e}\) values defines a genotype-conditioned phenotype whose
environmental explanation is evaluated prospectively. PC1 is used only as a statistical axis for defining continuous genomic
variation. It is not interpreted as a mechanistic genotype class, biological
pathway, or causal genomic effect.

\subsection{Explanation construction and prospective transport}
\label{sec:method_transport}

For each resolution $r$, the explanation is a signed 16-dimensional
vector over the same prespecified environmental concepts,
$\mathbf b^{(r)}=(b^{(r)}_1,\ldots,b^{(r)}_{16})$.
The analysis combines directions estimated by standardized
environment-level regression with nonnegative concept-usage
weights derived from an XGBoost/TreeSHAP model fitted to the
corresponding scientific target. The regression component
determines coordinate direction, while the weighting component
modulates the relative contribution of each concept.
This distinction matters: the signed vector should not be
interpreted as a direct average of signed TreeSHAP attributions.
The concept definitions, ordering, and evaluation procedure
are fixed across reference and target contexts within each split. For target year $t$, the reference explanation
$\mathbf b^{(r)}_{\mathrm{ref},t}$ is constructed from historical
environments and evaluated against the independently estimated
target-year explanation $\mathbf b^{(r)}_{\mathrm{target},t}$.
The primary transport statistic is cosine alignment,
\[
A_{\cos}^{(r)}(t)=
\frac{
\mathbf b^{(r)}_{\mathrm{ref},t}\cdot
\mathbf b^{(r)}_{\mathrm{target},t}
}{
\|\mathbf b^{(r)}_{\mathrm{ref},t}\|_2
\|\mathbf b^{(r)}_{\mathrm{target},t}\|_2}.
\]
Importantly, estimating the future-year explanation is a
\emph{retrospective audit} requiring observations from that year;
the target-year observations never enter the historical
reference explanation. Larger values indicate preservation
of direction rather than a causal mechanism. Concept-level
sign agreement is a secondary descriptive measure. For the confirmatory main-effect forecast, uncertainty is
summarized with $B=200$ balanced resamples of environment-level
rows using seed 42. The reported 80\% bootstrap percentile interval
is the empirical 10th--90th percentile interval of the
resampled forecasts and may be asymmetric.

\subsection{Measurement reliability and within-context reproducibility}
\label{sec:method_reliability}

Poor prospective transport can arise because the underlying scientific target is
itself difficult to estimate or because an estimable target acquires a different
environmental explanation in a future context. We therefore measure target
reliability independently of explanation transport. Within each environment, hybrids are randomly divided into two halves and
\(\beta_{G,e}\) is estimated independently in each half. Split-half reliability,
\(r_{\mathrm{half}}\), is the Pearson correlation between the two sets of
environment-level slopes. The corresponding full-sample reliability estimate is
obtained using the Spearman--Brown correction:
\[
r_{\mathrm{full}}
=
\frac{2r_{\mathrm{half}}}
{1+r_{\mathrm{half}}}.
\]
Reliability is interpreted as evidence that the scientific target is measurable,
rather than as a numerical benchmark or upper bound for prospective explanation
transport. We also assess whether low cross-year alignment could reflect unstable
construction of the G$\times$E explanation itself. Within a fixed context, we
independently resample the data used to construct the G$\times$E explanation and
recompute the explanation. Cosine similarity between these independently
resampled explanations measures within-context reproducibility. Unlike
prospective transport, both explanations in this comparison come from the same
context.

\subsection{Locked 2024 prospective evaluation}
\label{sec:method_locked}

Before the final successful 2024 scientific evaluation, the principal scientific
definitions were frozen: the 16 environmental concepts, the continuous
PC1-based G$\times$E target, the prospective transport metrics, the \(B=200\)
resampling procedure, the prespecified heat-direction control, and the expected
directions of all 16 main-effect concepts. Historical development supported a quantitative main-effect forecast, which was
fixed at
\[
A_{\cos}^{\mathrm{main}}(2024)=0.831,
\qquad
\mathrm{I}_{80}^{\mathrm{boot}}=[0.755,0.942].
\]
No numerical G$\times$E point forecast was specified because historical
interaction-level transport was too heterogeneous to support a defensible
prospective point prediction. The flowering/anthesis heat concept was
prespecified as a heat-direction control. An initial 2024 run failed before producing any preregistered scientific metric
because of a software compatibility issue. The repair did not change the frozen
scientific estimands, concepts, metrics, forecast, expected directions, or
resampling procedure.

\subsection{Secondary tests of alternative explanations}
\label{sec:method_secondary}

All analyses in this subsection were introduced after the confirmatory design
and are therefore reported as \emph{secondary/post-hoc}.

\paragraph{Validation-protocol comparison.}
We test whether apparent explanation reproducibility depends on the way
validation data are constructed. Three protocols are compared:
\textsc{Random-Env}, which randomly partitions historical environments while
ignoring calendar time; \textsc{Leave-Year-Out}, which holds out an entire year
but permits both earlier and later years in the reference set; and
\textsc{Strict-Forward}, which uses only past years to explain the subsequent
future year. Differences between the non-temporal protocols and strict-forward
evaluation are described as \emph{validation optimism gaps}, rather than
statistical bias, because they do not represent perfectly identical estimands.

\paragraph{Cross-method explanation robustness.}
To test whether the main-versus-G$\times$E hierarchy is specific to the
nonlinear XGBoost/SHAP pipeline, we fit a standardized Ridge regression using
the same 16 environmental concepts, treating its signed standardized
coefficients as the explanation vector without assuming numerical equivalence
to SHAP explanations.

\paragraph{Secondary-phenotype generalization.}
Finally, we test whether reduced interaction-level transport is an inevitable
statistical consequence of conditioning on genotype. A secondary phenotype,
Grain Moisture, was selected using a prespecified coverage rule before
its transport results were examined. It is directly observed, available across
the complete historical period, and analyzed with the same environmental concept
space and reference-derived genomic PC1 construction. The biological
concept-direction preregistration for grain yield does not apply to Grain Moisture.

\section{Results}
\label{sec:results}

\subsection{Yield explanations separate into coarse and interaction-level
prospective regimes}
\label{sec:results_primary}

We first evaluated whether explanations constructed from historical environments
retained their structure in the subsequent year. For grain yield, the two
scientific resolutions occupy markedly different prospective regimes. Across
the five historical target years, environment-level explanations achieved a
median cosine similarity of \(0.831\), whereas continuous G$\times$E
explanations achieved a median of \(0.295\)
(Table~\ref{tab:primary_results}). The corresponding median gap was \(0.536\). This difference reflects heterogeneity rather than a universal failure of
interaction-level explanations. In 2019, the G$\times$E explanation transported
strongly (\(A_{\cos}=0.822\)); in 2020, the corresponding cosine was only
\(0.006\). By contrast, main-effect transport remained high across all five
historical targets, ranging from \(0.736\) to \(0.962\). The locked 2024 evaluation was consistent with the historical ordering. The
observed main-effect cosine was \(0.811\), while the continuous G$\times$E
cosine was \(0.271\), yielding a main-to-interaction gap of \(0.540\). Only the
main-effect result was evaluated against a numerical forecast fixed before the
successful scientific evaluation: the expected cosine was \(0.831\) with an
\(80\%\) bootstrap percentile interval of \([0.755,0.942]\). The observed value fell
inside this interval, with an absolute forecast error of \(0.020\). No
numerical G$\times$E point forecast had been specified because the historical
interaction-level regime was too heterogeneous to justify one.

\begin{table}[h]
\centering
\caption{\textbf{Primary prospective evidence for grain yield.}
Historical values summarize the five strictly forward evaluations
(2019--2023). The 2024 main-effect bootstrap percentile interval was locked before the
successful scientific evaluation. No numerical G$\times$E forecast was
specified.}
\label{tab:primary_results}
\small
\setlength{\tabcolsep}{5pt}
\begin{tabular}{lcc}
\toprule
\textbf{Quantity} & \textbf{Historical} & \textbf{2024} \\
\midrule
Main-effect cosine
    & \(0.831\) median
    & \(0.811\) \\
G$\times$E cosine
    & \(0.295\) median
    & \(0.271\) \\
Main--G$\times$E gap
    & \(0.536\)
    & \(0.540\) \\
Main forecast
    & --
    & \(0.831\) \\
Main \(80\%\) bootstrap percentile interval
    & --
    & \([0.755,0.942]\) \\
Main forecast outcome
    & --
    & inside interval \\
Prespecified heat-direction control
    & --
    & PASS \\
Frozen concept directions
    & --
    & \(14/16\) \\
\bottomrule
\end{tabular}
\end{table}

The concept-level checks provide an additional prospective constraint on the
2024 result. The preregistered flowering/anthesis heat control retained its
expected negative direction, changing from a historical reference explanation
coordinate of \(-0.791\) to \(-0.989\) in 2024. Fourteen of the 16 frozen
main-effect concept directions matched their prespecified signs. The two
exceptions were grain-filling water (expected negative, observed \(+0.892\))
and midseason soil-water storage (expected negative, observed \(+0.247\)). We
retain these failures rather than filtering them from the prospective evaluation.

\subsection{Measurement reliability and within-context reproducibility do not
explain the interaction-level transport loss}
\label{sec:results_reliability}

One possible explanation for the lower G$\times$E transport is that the
environment-specific genomic-response phenotype is simply too noisy to
estimate. The reliability audit does not support this explanation as
sufficient. For the continuous \(\beta_G\) phenotype, the canonical split-half
reliability was \(r_{\mathrm{half}}=0.554\), corresponding to
\(r_{\mathrm{full}}=0.713\) after Spearman--Brown correction. Despite this
measurable target reliability, the historical median prospective G$\times$E
cosine was \(0.295\). The year-wise contrast is more informative than the aggregate value. In 2019,
relatively modest split-half reliability (\(0.352\)) coexisted with strong
prospective G$\times$E transport (\(0.822\)). In 2020, reliability increased to
\(0.798\), yet future explanation alignment fell to \(0.006\). Measurement
precision and future explanation alignment therefore do not exhibit a simple
monotonic relationship. To distinguish cross-year transport from within-context estimation variability,
we independently resampled the data used to construct each G$\times$E
explanation. Across the 2020--2023 transitions, within-context reproducibility
ranged from \(0.77\) to \(0.95\), whereas prospective alignment with the
subsequent year ranged from \(0.01\) to \(0.36\). The year-specific contrasts
were \(0.87\)--\(0.95\) versus \(0.01\) (2020), \(0.86\)--\(0.91\) versus
\(0.36\) (2021), \(0.87\)--\(0.91\) versus \(0.30\) (2022), and
\(0.77\)--\(0.81\) versus \(0.18\) (2023). Thus, the reduced future-year
alignment in these transitions is not attributable solely to instability in
estimating the environment-specific genomic-response target.

\subsection{Non-temporal validation selectively overstates interaction
reproducibility}
\label{sec:results_protocol}

We next tested whether the apparent reproducibility of an explanation depends
on the validation protocol. This comparison reveals an asymmetric effect
(Table~\ref{tab:stress_tests}). For the environmental main effect, randomly
partitioning environments produced a median cosine of \(0.793\), slightly
\emph{below} the strict-forward value of \(0.831\), corresponding to an
optimism gap of \(-0.039\). The coarse-resolution result therefore does not
benefit from random non-temporal validation. For G$\times$E, the pattern is reversed. Random-environment validation yielded
a cosine of \(0.512\), compared with \(0.295\) under strict-forward evaluation,
an optimism gap of \(+0.217\). Leave-year-out validation showed the same
direction, with a G$\times$E optimism gap of \(+0.154\). Thus, protocols that
do not enforce a past-to-future ordering make the interaction-level explanation
appear substantially more reproducible than it is under the scientific target
of interest: explaining a genuinely future year. We describe these differences as \emph{validation optimism gaps}, rather than
formal bias, because the protocols are closely matched but do not correspond to
perfectly identical estimands. The result nevertheless demonstrates that
evaluation design can selectively alter the apparent validity of fine-resolution
explanations.

\subsection{Interaction-level transport is phenotype-dependent}
\label{sec:results_phenotype}

The large yield main-to-G$\times$E gap raises a broader question: is poor
interaction-level transport an inevitable statistical consequence of
conditioning an explanation on genotype? A coverage-locked secondary analysis
of Grain Moisture argues against that interpretation. For Grain Moisture, the median main-effect cosine was \(0.907\), while the
median G$\times$E cosine remained high at \(0.859\). Their gap of medians was
only \(0.048\), compared with \(0.536\) for yield
(Table~\ref{tab:stress_tests}). Measurement reliability was also high at both
levels: \(0.998\) for the environment mean and \(0.843\) for the PC1-slope
phenotype. Yield therefore shows a large coarse-to-interaction gap, whereas Grain
Moisture retains strong transport at both levels. This contrasting
phenotype is a crucial boundary case: the audit can identify strongly
transporting genotype-conditioned explanations, so reduced transport
is not a universal consequence of finer scientific resolution.
The yield-specific biological sign preregistration is not transferred
to Grain Moisture.

\subsection{The main-to-interaction hierarchy persists under an independent
explanation model}
\label{sec:results_method}

Finally, we tested whether the primary yield hierarchy is specific to the
XGBoost/SHAP explanation family. Under the primary nonlinear pipeline, median
transport was \(0.831\) for the main effect and \(0.295\) for G$\times$E, a gap
of \(0.536\). Under standardized Ridge regression, the corresponding values
were lower in absolute terms---\(0.588\) and \(0.227\)---but their ordering was
preserved, with a gap of \(0.361\). The main-to-interaction ordering therefore persists under an independent linear
explanation family, although absolute transport remains method-dependent.

\begin{table}[h]
\centering
\caption{\textbf{Secondary tests of alternative explanations.}
All analyses in this table are secondary/post-hoc and leave the locked 2024
confirmatory analysis unchanged. Values are prospective cosine similarities
unless otherwise noted.}
\label{tab:stress_tests}
\small
\setlength{\tabcolsep}{4.5pt}
\begin{tabular}{lccc}
\toprule
\textbf{Stress test} &
\textbf{Coarse/Main} &
\textbf{Fine/G$\times$E} &
\textbf{Difference} \\
\midrule
Yield, XGBoost/SHAP
    & \(0.831\)
    & \(0.295\)
    & \(0.536\) \\
Yield, Ridge
    & \(0.588\)
    & \(0.227\)
    & \(0.361\) \\
Grain Moisture
    & \(0.907\)
    & \(0.859\)
    & \(0.048\) \\
Random-Env validation
    & \(0.793\)
    & \(0.512\)
    & -- \\
Strict-Forward validation
    & \(0.831\)
    & \(0.295\)
    & -- \\
Validation optimism gap
    & \(-0.039\)
    & \(+0.217\)
    & -- \\
\bottomrule
\end{tabular}
\end{table}

\subsection{Post-hoc sensitivity to weighting and similarity metrics}
\label{sec:results_new_sensitivity}

With signed environment-level regression coefficients fixed, varying
the XGBoost/SHAP weighting seed over five runs preserved the main
$>$ G$\times$E ordering in every run for 2020--2024 (but not 2019).
In 2024 the resulting mean $\pm$ SD cosines were $0.809\pm0.008$
and $0.265\pm0.008$, respectively. This assesses the weighting
stage \emph{only}, not full-pipeline stability.
For the full-data 2024 explanations, unweighted cosine
($0.789$ vs.\ $0.145$), absolute-importance Spearman correlation
($0.388$ vs.\ $-0.697$), top-five Jaccard overlap
($0.429$ vs.\ $0.000$), and SHAP-weighted sign agreement
($0.831$ vs.\ $0.549$) each retained the main $>$ G$\times$E
ordering. Full results appear in Tables~\ref{tab:seed_robustness}
and~\ref{tab:metric_robustness_2024} in
Appendix~\ref{app:seed_metric_sensitivity}; neither
check changes the locked 2024 primary evaluation.

\subsection{Result synthesis}
\label{sec:results_synthesis}

The locked 2024 evaluation was consistent with
the historical yield main-to-G$\times$E ordering, while only the main-effect
cosine received a numerical prospective forecast. Non-temporal
validation selectively inflates apparent interaction-level reproducibility.
The same transport gap is not inevitable: Grain Moisture maintains
strong transport at both coarse and genotype-conditioned levels, while the
main-to-interaction ordering for yield persists under Ridge.
The observed transport gap does not by itself distinguish
contextual association changes from measurement or estimation effects.
Prospective scientific validity depends
jointly on the \emph{resolution of the claim}, the \emph{phenotype being
explained}, and the \emph{evaluation protocol}.

\section{Discussion and Limitations}
\label{sec:discussion}

Our findings distinguish \emph{target measurability},
\emph{within-context reproducibility}, and
\emph{prospective explanation transportability}.
For grain yield, genotype-conditioned explanations
showed substantially lower and more heterogeneous
future-year alignment than main-effect explanations,
despite measurable targets and high within-context
reproducibility.

\paragraph{Robustness and scope.}
Alternative similarity metrics preserved the 2024
main-to-interaction ordering, indicating that the
contrast is not specific to weighted cosine.
The five-seed analysis showed robustness to
XGBoost/SHAP weighting randomness, but does not
establish stability of the entire estimation pipeline.

\paragraph{Interpretation.}
The observed transport gap may reflect covariate
drift, model refitting, or estimation variability
rather than a uniquely identifiable biological
mechanism. Nevertheless, prospective explanation
transport remains a distinct validation question.
Grain Moisture provides a counterexample, with
strong transport at both explanation resolutions.

\paragraph{Limitations.}
The study uses one maize resource, five historical
future-year evaluations, and correlated environmental
concepts. Genomic PC1 is not a mechanistic axis,
and small environment samples constrain precision.
All findings are observational rather than causal.

\section*{Responsible Use Statement}
Our results are observational and should not be
interpreted as causal evidence. They should be
validated prospectively before use in new settings.

\section{Conclusion}
\label{sec:conclusion}

Reliable measurement and within-context explanation
reproducibility do not necessarily establish prospective
explanation transportability. Across multi-year maize
field trials, main-effect explanations for grain yield
showed substantially stronger future-year alignment
than genotype-conditioned explanations, despite high
within-context reproducibility. The locked 2024
evaluation reinforced this contrast, while non-temporal
validation overstated interaction-level transport.
Additional analyses using Ridge explanations and
alternative similarity metrics supported the overall
ordering, whereas Grain Moisture demonstrated that
strong transport at both resolutions is possible.
These findings do not identify a unique mechanism
behind transport failures. Instead, they establish
prospective explanation auditing as a complementary
validation requirement: explanations intended to
support reproducible scientific claims should be
evaluated in genuinely future contexts, beyond
predictive accuracy, measurement reliability, or
within-context stability.

\bibliographystyle{plainnat}
\bibliography{references}

\clearpage
\appendix
\section{Additional sensitivity checks on the explanation contrast}
\label{app:seed_metric_sensitivity}

All analyses in this appendix were introduced after the locked primary
evaluation. Numbers are transcribed from the reported secondary-analysis
outputs; these tables do not constitute an independent reproduction of
the underlying computations.

\paragraph{Sensitivity to SHAP-weight seed.}
The environment-level standardized regression coefficients were held
fixed while the XGBoost-derived usage/SHAP weighting stage was varied
over five seeds. Thus, the experiment does not assess uncertainty from
re-estimating the response phenotype or the regression coefficients.
The across-seed main-versus-G$\times$E ordering is preserved for each
target year from 2020 through 2024, but reversed for 2019.

\begin{table}[h]
\centering
\caption{\textbf{Post-hoc SHAP-weight seed sensitivity.}
Mean $\pm$ standard deviation of weighted cosine across five XGBoost
seeds, with the signed regression coefficients held fixed.
The last column counts seeds in which Main $>$ G$\times$E.}
\label{tab:seed_robustness}
\small
\setlength{\tabcolsep}{5pt}
\begin{tabular}{lccc}
\toprule
Target year & Main (mean $\pm$ SD) & G$\times$E (mean $\pm$ SD) & Main $>$ G$\times$E \\
\midrule
2019 & $0.777\pm0.016$ & $0.870\pm0.006$ & $0/5$ \\
2020 & $0.826\pm0.011$ & $0.029\pm0.026$ & $5/5$ \\
2021 & $0.741\pm0.009$ & $0.390\pm0.022$ & $5/5$ \\
2022 & $0.920\pm0.003$ & $0.308\pm0.019$ & $5/5$ \\
2023 & $0.964\pm0.000$ & $0.212\pm0.022$ & $5/5$ \\
2024 & $0.809\pm0.008$ & $0.265\pm0.008$ & $5/5$ \\
\bottomrule
\end{tabular}
\end{table}

\paragraph{Alternative similarity summaries.}
The following post-hoc metrics were evaluated on full-data 2024
explanation objects, rather than on the original $B=200$ resampling
median (the original locked cosine values remain $0.811$ and $0.271$).
Unweighted cosine uses the unweighted standardized coefficients;
Spearman and top-five Jaccard use absolute-importance rankings;
SHAP-weighted sign agreement summarizes consistency of coordinate
directions under the weighting convention. These metrics characterize
different aspects of the explanation, and should not be interpreted
as four replications of the same estimand.

\begin{table}[h]
\centering
\caption{\textbf{Post-hoc alternative metrics in 2024.}
Full-data estimates; main versus genotype-conditioned response.}
\label{tab:metric_robustness_2024}
\small
\setlength{\tabcolsep}{7pt}
\begin{tabular}{lcc}
\toprule
Metric & Main & G$\times$E \\
\midrule
Unweighted cosine & $0.789$ & $0.145$ \\
Spearman correlation (absolute importance) & $0.388$ & $-0.697$ \\
Top-5 Jaccard (absolute importance) & $0.429$ & $0.000$ \\
SHAP-weighted sign agreement & $0.831$ & $0.549$ \\
\bottomrule
\end{tabular}
\end{table}

The main-versus-G$\times$E ordering in 2024 is consistent across
these summaries, but the distinct metrics capture different notions
of agreement. In particular, the results do not identify why the
underlying explanations differ across years.

\end{document}